\documentclass[10pt,twocolumn,letterpaper]{article}

\usepackage{iccv}
\usepackage{times}
\usepackage{epsfig}
\usepackage{graphicx}
\usepackage{amsmath}
\usepackage{amssymb}

\usepackage[table]{xcolor}
\usepackage{booktabs,arydshln}
\usepackage{mathtools}
\usepackage{tabularx}
\usepackage{enumitem}
\usepackage{multirow}
\usepackage{subcaption}
\usepackage{wrapfig}
\usepackage{xcolor}
\usepackage{soul}
\usepackage{multirow}
\usepackage{subcaption}
\usepackage{wrapfig}
\usepackage{tabularx}
\usepackage{booktabs,arydshln}
\usepackage{makecell}
\usepackage{wrapfig,lipsum,booktabs}

\usepackage{amsmath,amsfonts,bm}

\def\eqref#1{Eq.~\ref{#1}}

\def\1{\bm{1}}

\DeclareMathAlphabet{\mathsfit}{\encodingdefault}{\sfdefault}{m}{sl}
\SetMathAlphabet{\mathsfit}{bold}{\encodingdefault}{\sfdefault}{bx}{n}

\makeatother

\definecolor{MyDarkBlue}{rgb}{0,0.08,1}
\definecolor{MyDarkGreen}{rgb}{0.02,0.6,0.02}
\definecolor{MyDarkRed}{rgb}{0.8,0.02,0.02}
\definecolor{MyDarkOrange}{rgb}{0.40,0.2,0.02}
\definecolor{MyPurple}{RGB}{111,0,255}
\definecolor{MyRed}{rgb}{1.0,0.0,0.0}
\definecolor{MyGold}{rgb}{0.75,0.6,0.12}
\definecolor{MyDarkgray}{rgb}{0.66, 0.66, 0.66}

\newcommand{\xhdr}[1]{\vspace{0pt} \noindent {\textbf{#1}}}

\usepackage{pifont}
\newcommand{\cmark}{\ding{51}}%
\newcommand{\xmark}{\ding{55}}%

\usepackage[pagebackref=true,breaklinks=true,colorlinks,bookmarks=false]{hyperref}

\iccvfinalcopy 

\def\iccvPaperID{****} 

\newcommand{\oursloss}{\textbf{\texttt{TACo}}}
\newcommand{\ourslossfull}{\textbf{\texttt{T}}oken-\textbf{\texttt{A}}ware \textbf{\texttt{C}}ascade  c\textbf{\texttt{o}}ntrastive learning (\textbf{\texttt{TACo}})}

\begin{document}

\title{TACo: Token-aware Cascade Contrastive Learning for Video-Text Alignment}

\author{Jianwei Yang\\
Microsoft Research\\
{\tt\small jianwyan@microsoft.com}
\and
Yonatan Bisk\\
Carnegie Mellon University\\
{\tt\small ybisk@cs.cmu.edu}
\and
Jianfeng Gao\\
Microsoft Research\\
{\tt\small jfgao@microsoft.com}
}

\maketitle
\ificcvfinal\thispagestyle{empty}\fi

\begin{abstract}
Contrastive learning has been widely used to train transformer-based vision-language models for video-text alignment and multi-modal representation learning. This paper presents a new algorithm called \ourslossfull{} 
that improves contrastive learning using two novel techniques. 
The first is the token-aware contrastive loss which is computed by taking into account the syntactic classes of words. This is motivated by the observation that for a video-text pair, the content words in the text, such as nouns and verbs, are more likely to be aligned with the visual contents in the video than the function words.
Second, a cascade sampling method is applied to generate a small set of hard negative examples for efficient loss estimation for multi-modal fusion layers. To validate the effectiveness of \oursloss, in our experiments we finetune pretrained models for a set of downstream tasks including text-video retrieval (YouCook2, MSR-VTT and ActivityNet), video action step localization (CrossTask), video action segmentation (COIN). The results show that our models attain consistent improvements across different experimental settings over previous methods, setting new state-of-the-art on three public text-video retrieval benchmarks of YouCook2, MSR-VTT and ActivityNet. 
\end{abstract}

\section{Introduction}
Aligning or grounding language to videos is a challenging topic in the context of vision-language (VL) research as it requires the model to understand contents, dynamics, and causality presented in videos~\cite{Bisk2020}. Inspired by the success of BERT~\cite{Devlin2018} in natural language processing, there is a growing interest in applying transformer-based multi-modal models for video-text alignment and representation learning~\cite{videobert, sun2019learning, Zhu2020, Luo2020, gabeur2020multi, li2020hero}. 
These models are typically pretrained on large amounts of noisy video-text pairs using contrastive learning~\cite{Miech2019, Miech2020}, and then applied in a zero-shot manner or finetuned for various downstream tasks, such as text-video retrieval~\cite{Xu2016c}, video action step localization~\cite{zhukov2019cross}, video action segmentation~\cite{tang2019coin}, video question answering~\cite{tapaswi2016movieqa,lei2018tvqa} and video captioning~\cite{Zhou2017a}.

\begin{figure}[t]
    \centering
    \includegraphics[width=0.95\linewidth]{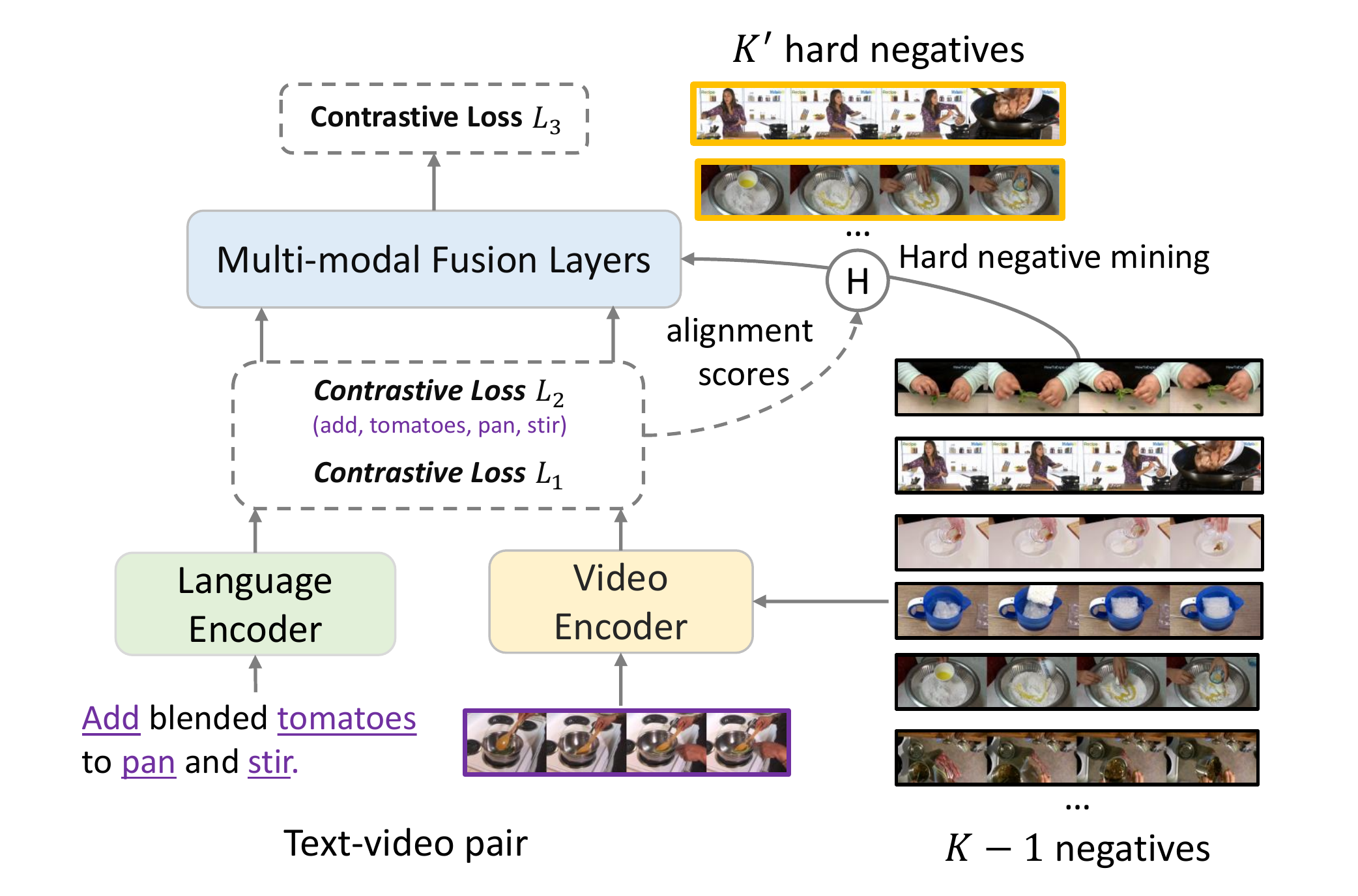}
    \vspace{-7pt}
    \caption{The proposed token-aware cascade contrastive learning pipeline. We compute three contrastive losses: 1) sentence-level loss $L_1$ over all negative examples; 2) token-level loss $L_2$ on content words (noun, verb) over all negative examples; 3) sentence-level loss $L_3$ over hard negative examples sampled based on $L_1$ and $L_2$ online.}
    \label{fig:teaser_figure}
\vspace{-15pt}
\end{figure}

In this paper, we present a new variant of contrastive learning,  
\ourslossfull{}
to improve the video-text alignment for both large-scale pretraining and downstream specific tasks.
As the name indicates, \oursloss{} makes two modifications to the conventional contrastive learning used in video-language domain. The first is the token-aware contrastive loss which is computed by taking into account the syntactic classes of words. This is motivated by the observation that, given a video and its corresponding text, content words, such as nouns and verbs, are more likely than function words to be aligned with (or grounded to) visual contents in the video. Conventional contrastive learning typically compute the loss 
after aggregating over all the words in the text and frames in the video (loss $L_1$ or $L_3$ in Fig.~\ref{fig:teaser_figure}). In contrast, the token-aware contrastive loss is computed using only a subset of words whose syntactic classes belong to a pre-defined set (\textit{e.g.}, nouns and verbs), which forces the grounding of individual words to the video (loss $L2$). For example, we pay particular attention to the words ``add'', ``tomatos'', ``pan'' and ``stir'' in Fig.~\ref{fig:teaser_figure}.

The second technique we introduce is a cascade sampling method to find a small set of hard negative examples for training the multi-modal fusion layers. Consider a batch of $K$ video-text pairs. For each of the video-text pairs, the ideal case is that we use the remaining $K-1$ negative videos or texts to compute the contrastive loss after multi-modal fusion. However, the cost of computing the contrastive loss quickly becomes prohibitive when it is coupled with multi-modal fusion layers, considering its high complexity $O(K^2 \times L^2)$ where $L$ is total number of visual and textual tokens. A conventional way to address this is using random sampling to select a small subset of negative pairs. In this paper, instead of random sampling, we propose a cascade sampling method as shown in the top-right of Fig.~\ref{fig:teaser_figure} to efficiently select a small set of hard negative examples \emph{on the fly} during training. It leverages the video-text alignment scores computed in $L_1$ and $L_2$ before multi-modal fusion layers, and helps to learn the multi-modal fusion layers more effectively without any extra overhead.

We perform a comprehensive empirical study to validate the effectiveness of \oursloss{} in both pretraining and dataset-specific scenarios. We apply \oursloss{} and different variants of contrastive losses to train or pretrain and finetune on various downstream tasks including text-video retrieval (YouCook2, MSR-VTT and ActivityNet)~\cite{Zhou2017a, Xu2016c,caba2015activitynet}, video action step localization (CrossTask)~\cite{zhukov2019cross} and action segmentation (COIN)~\cite{tang2019coin}. Our results show that \oursloss{} improves the text-video retrieval performance over current state-of-the-art across three benchmarks. Furthermore, the learned multi-modal representation and video representation can be effectively transferred to CrossTask and COIN, and achieve better or comparable performance to current state-of-the-art methods.

\section{Related work}
\label{Sec:relatedwork}
\xhdr{Video-language pretraining}. Realistic application scenarios around videos have prompted emergence of various video-language tasks, such as text-video retrieval~\cite{lin2014microsoft,yu2017end,yu2018joint}, video question answering~\cite{jang2017tgif, lei2018tvqa}, video captioning~\cite{yu2016video,zhou2018end}, \textit{etc}. Inspired by the success of BERT for large-scale pretraining in language domain~\cite{Devlin2018}, transformers have been employed in the video-language domain~\cite{videobert,Zhu2020,Luo2020, li2020hero} as well as image-language domain~\cite{Lxmert2019,Lu2019,Zhou2019a,li2020oscar}. Combined with large scale datasets, \textit{e.g.} Howto100M~\cite{Miech2019} this approach has proven to be effective on various downstream tasks. Depending on the tasks of interest, some approaches train a multi-modal transformer using a combination of multiple losses including video-text alignment~\cite{videobert,Zhu2020,Luo2020,li2020hero}, 
masked token (words/frames/objects) prediction~\cite{videobert,Zhu2020,Luo2020}, 
and frame order prediction~\cite{li2020hero}, \textit{etc}. 
Some other approaches exploited various contrastive learning techniques to directly optimize the feature space without multi-modal fusion~\cite{Miech2019, Miech2020, liu2019use, gabeur2020multi}. In most of previous works, these two approaches were explored separately. Very recently, an updated version of~\cite{Luo2020} used two independent alignment losses before and after multi-modal fusion in a single framework. In this paper, however, these two losses cooperate closely with each other during training in that the earlier stage helps to discover the hard negatives while the multi-modal layers with more capacity help to tackle those hard samples particularly.

\xhdr{Video-text alignment}. Aligning videos to text requires the model to understand motion and temporal coherence. Some works have relied on attention mechanisms to extract key information from videos \cite{torabi2016learning, yu2017end}, while others preserve visual information by composing pairwise joint representation using 3D tensors~\cite{yu2018joint} or use multi-level video encoders to separately encode the spatial and temporal cues~\cite{dong2019dual}. These models usually rely on a rank or margin loss to learn the correct alignment for video-text pairs. 
Another line of work learns fine-grained or hierarchical alignment between videos and texts~\cite{zhang2018cross,wray2019fine,chen2020fine}. In~\cite{wray2019fine}, the authors proposed a fine-grained alignment by extracting the nouns and verbs from action phrase in a sentence and projecting them into a shared space with videos. Alternatively, the authors in~\cite{chen2020fine} extract a hierarchical semantic graph and apply graph reasoning to achieve the alignment at different levels. Similar ideas have been also proposed in the image-text alignment by decomposing the images and texts into sub-tokens~\cite{lee2018stacked,wu2019unified}. Thus far, it has not been studied how these task-specific architectures can be integrated into large-scale pretraining. In this paper, we are the first to propose a simple yet effective token-aware contrastive loss for fine-grained alignment for pretraining and downstream tasks.

\xhdr{Negative sampling}. Key to efficient contrastive training is a good source of negative examples. 
Most of current approaches use random sampling strategies for training video-text alignment~\cite{Zhu2020,Luo2020}. However, in the domain of image-text retrieval, a few works tried hard negative sampling to choose the hardest negatives for training. In~\cite{appalaraju2017image,faghri2017vse++}, the authors computed the alignment scores for all image-text pairs in a mini-batch and use the hardest negative sample to compute the marginal loss. However, this strategy can only be applied without multi-modal fusion. In those models which have multi-modal fusion layers for better representations~\cite{Lu2019,chen2020uniter}, the authors instead compute the matching score offline and then use it to sample hard negatives for finetuning image-text retrieval model, which however is difficult for large-scale pretraining due to the high computational cost. In this paper, our cascade hard negative mining is particularly designed to address these issues as we efficiently select the hard negative samples online before multi-modal fusion and send them to the fusion layers for computing the loss. As we will show in our experiments, this technique can be seamlessly applied to both large-scale pretraining and downstream tasks.


\section{Method}

\subsection{Framework}
\label{subsec:framework}
As depicted in Fig.~\ref{fig:teaser_figure}, our model has three components: 

\xhdr{Video encoding module $f_{\theta_v}$}. It is implemented by a stack of self-attention layers parameterized by $\theta_v$. Here, we assume the input video features have been already extracted using some pre-trained models such as 2D CNN (\textit{e.g.}, ResNet~\cite{He2015}) or 3D CNN (\textit{e.g.}, I3D~\cite{Carreira:2017}, S3D~\cite{xie2018rethinking}). Given the input video embeddings, video encoder starts with a linear layer to project them to the same dimension $d$ as following self-attention layers. We denote the output of our video encoder for a video clip by a sequence of $m$ features, $\bm{x} =\{x^1, ...,x^m\} \in \mathbb{R}^{m \times d}$. The number of features $m$ depends on the choice of sampling frame rate and the video feature extractor, which we will discuss in Sec.~\ref{Sec:experiment}. 

\xhdr{Language encoding module $f_{\theta_t}$}. We use pretrained tokenizer~\cite{Wolf2019HuggingFacesTS} and BERT~\cite{Devlin2018} to tokenize the input texts and extract textual features, respectively. Given a raw sentence, we append a ``[CLS]'' and ``[SEP]'' to the beginning and end, respectively. At the top, we can obtain a sequence of $n$ textual features $\bm{y} = \{y^1,...,y^n\} \in \mathcal{R}^{n \times d}$. We ensure the output feature dimension of video encoder to be identical to that of language encoder. During training, we update the parameters $\theta_t$ in our language encoder to adapt to the texts in specific domain, \textit{e.g.,} cooking instructions in YouCook2~\cite{Zhou2017a}.

\xhdr{Multi-modal fusion module $f_{\theta_m}$}. It also consists of self-attention layers with learnable parameters $\theta_m$. It takes video features $\bm{x} \in \mathbb{R}^{m \times d}$ and text features $\bm{y} \in \mathbb{R}^{n \times d}$ from two separate modalities as inputs and output the $(m+n)$ features $\bm{z} = \{z_1,...,z_{(m+n)}\} \in \mathcal{R}^{(m+n) \times d}$. To help it to distinguish the video and language tokens, we use a token type embedding layer to learn two embeddings and add them to the visual and textual tokens, separately. Similar to original Transformer~\cite{Vaswani2017}, we include a positional embedding layer to encode the absolute token positions in the input sequence.

The above three components comprise our video-text alignment model which is then trained with the proposed token-aware cascade contrastive loss. We start with a brief review of conventional contrastive learning and then introduce the proposed technique.

\subsection{Contrastive learning: a revisit}

Given a set of $N$ video-text pairs $\{(v_i, t_i)\}_{i=1}^N$, our goal is to learn an optimal scoring function $s$ such that paired video and text $(v_i, t_i)$ have higher scores than all the other unmatched pairs $(v_j, t_k), j\neq k$. From the probabilistic perspective, aligning $v_i$ to $t_i$ is equivalent to maximizing the conditional probability $p(v_i|t_i)$ while minimizing the probability for all negative pairs $p(v_j|t_i), j\neq i$. According to~\cite{gutmann2010noise,mnih2012fast}, $p(v_j|t_i)$ can be approximated by:
\begin{equation}
\small
    p({v}_j | {t}_i) \sim \frac{\exp^{s({v}_j, {t}_i)}}{\sum_{k=1}^{N} \exp^{s({v}_k, {t}_i)}}
    \label{joint_probability}
\end{equation}
where $s(v, t)$ is the alignment score between $v$ and $t$; the denominator is a sum over all possible videos, which is a partition function for normalization. Adding cross-entropy loss on $p({v}_j | {t}_i)$, we can then derive the NCE loss~\cite{gutmann2010noise}:
\begin{equation}
\small
\begin{split}
    L_{nce} &= \sum_{i=1}^{N} -\log p(v_i|t_i) \\ & \sim \sum_{i=1}^{N} -\log \left(\frac{\exp^{s({v}_i, {t}_i)}}{\exp^{s({v}_i, {t}_i)} + \sum_{k \neq i} \exp^{s({v}_k, {t}_i)}}\right)
\end{split}
\label{Eq:NCE}
\end{equation}

The denominator in \eqref{Eq:NCE} requires a sum over all videos in a dataset, which is intractable in practice. Therefore, we usually compute the NCE loss on a mini-batch of $K (K \ll N)$ video-text pairs sampled from the whole dataset. Ideally, we want to learn the parameters $\theta = \{\theta_v, \theta_t, \theta_m\}$ of the model to minimize the above NCE loss, such that $\Delta=s(v_i, t_i) - s(v_j, t_i)$ is maximized over all tuples $(t_i, v_i, v_j), j \neq i$. 
A number of previous works used the above formula for contrastive learning \cite{Miech2020, Zhu2020}. Meanwhile, there are some variants of computing contrastive loss in video-langauge representation learning. For example, \cite{li2020hero, gabeur2020multi} omits the denominator and incorporate a margin s.t. $ s(v_i, t_i) > s(v_j, t_i) + \delta, \forall j \neq i$ in a mini-batch. \cite{Luo2020} optimizes binary cross-entropy (BCE) by assigning $(v_i, t_i)$ a positive label (1) and other pairs a negative label (0). 

\subsection{\large \textbf{\texttt{TACo}}: our approach}
\label{sec:toco}

The way of using contrastive learning in previous works has two issues. First, the loss is computed at sentence-level by taking `[CLS]' token~\cite{gabeur2020multi} or the maximum over all tokens~\cite{Miech2020} in a sentence. Clearly, the content words (\textit{e.g.}, nouns, verbs) are more likely to align with the visual contents or concepts in the videos compared with function words (\textit{e.g.}, stop words). Second, the high computational cost in multi-modal fusion layers hinder the usage of large batch of negative samples, which however is essential to contrastive learning~\cite{Miech2020, he2020momentum, chen2020simple}. Motivated by these two issues, we introduce \oursloss{}, a simple yet effective method to improve the contrastive learning. We elaborate below how these contrastive losses are computed.

Given the $K$ video-text pairs $\{(v_i, t_i)\}_{i=1}^{K}$ in a mini-batch, we first use our video encoder $f_{\theta_v}$ and language encoder $f_{\theta_t}$ to obtain a batch of video features $X = \{x_1,...,x_K\} \in \mathcal{R}^{K \times m \times d}$ and text features $Y = \{y_1, ..., y_K\} \in \mathcal{R}^{K \times n \times d}$, respectively. Then, we average all tokens of a video clip $v_i$ to get $\bar{x}_i \in \mathcal{R}^{1 \times d}$, and take the first `[CLS]' token for each text $t_i$ to get $\bar{y}_i \in \mathcal{R}^{1 \times d}$. Based on $\bar{x}$ and $\bar{y}$, we compute the sentence-level contrastive loss:
\begin{equation}
\small
    L_1 = -\sum_{i=1}^{K} \log \left( \frac{\exp^{\bar{x}_i \cdot \bar{y}_i / \tau_1}}{\exp^{\bar{x}_i \cdot \bar{y}_i  / \tau_1} + \sum_{j \neq i}\exp^{\bar{x}_j \cdot \bar{y}_i / \tau_1}}\right)
    \label{EQ:contrastive_loss_ge}
\end{equation}
where $\tau_1$ is a scalar temperature parameter. In~\eqref{EQ:contrastive_loss_ge}, the computation is simply a number of dot-products between video and text features. Giving such efficiency, we can use all the $K-1$ negative samples in a mini-batch to compute the loss. Through this, we optimize $\theta_v$ and $\theta_t$ so as to project the video and text samples into an aligned feature space.


The `[CLS]' token and average of video tokens in~\eqref{EQ:contrastive_loss_ge} overlooks the differences across tokens and frames, and thus may not provide the pressure to push individual tokens (\textit{e.g.}, nouns and verbs) to ground on the specific video contents. To encourage correct alignment, in addition to the sentence-level loss, we introduce a token-level contrastive loss:
\begin{equation}
\small
    L_2 = -\sum_{i=1}^{K} \sum_{p \in \mathcal{P}_i} \log\left(\frac{\exp^{s(x_i, {y}^p_i) / \tau_2}}{\exp^{s(x_i, {y}^p_i) / \tau_2} + \sum\limits_{j \neq i} \exp^{s(x_j, {y}^p_i) / \tau_2} }\right) 
    \label{EQ:contrastive_loss_te}
\end{equation}
where $\tau_2$ is another scalar temperature parameter; $P_i$ is the indices of tokens of interest in $i$-th text, and $y_i^p$ is the $p$-th token embedding in $i$-th text. $s(\cdot)$ measures the similarity between video features and specific token embedding $y_i^p$. It first computes the dot-product between $y_i^p \in \mathcal{R}^{1 \times d}$ and all $m$ video tokens $x \in \mathcal{R}^{m \times d}$, and then take the maximum over $m$ scores to get the final alignment score. Through \eqref{EQ:contrastive_loss_te}, the model uses individual tokens as anchors to align with video, which is complementary to the sentence-level loss in \eqref{EQ:contrastive_loss_ge}. Similar to~\eqref{EQ:contrastive_loss_ge}, we can compute this token-level contrastive loss efficiently, and thus use all the $K-1$ negative samples. As a whole, these two losses are used to optimize $\theta_v$ and $\theta_t$ in a token-aware manner.

\xhdr{Token of interest}. In \eqref{EQ:contrastive_loss_te}, we need to decide which tokens should be included in $P_i$. In this paper, we heuristically select \textit{nouns} and \textit{verbs} as the targets considering they are more ``concrete'' in the videos. In practice, \textit{nouns} or \textit{verbs} usually have different discriminativeness even if they are all the same type. For example, ``man'' is a \textit{noun} but is less informative than ``gymnast''. To reflect this, we further assign different words with different weights by computing their inverse document frequency (idf) \cite{Jones72astatistical}. A higher idf means it is more unique across the corpus, and hence will weigh more when computing the token-level contrastive loss. Another practical issue for computing the loss is that the tokens are usually sub-words due to the BERT tokenizer. Hence, for all tokens that belongs to the same word, we will assign the same weights accordingly.

After computing the token-aware contrastive loss, we feed the features from separate modalities to multi-modal fusion layers to enable more interactions between them two. Similar to previous work~\cite{Zhu2020}, we take the feature corresponding to the ``[CLS]'' in the $(m+n)$ outputs. We regard this as the summary of two modalities and then compute the contrastive loss:
\begin{equation}
\small
    L_3 = -\sum_{i=1}^{K} \log \left( \frac{\exp^{w \cdot z^{cls}_{i,i}}}{\exp^{w \cdot z^{cls}_{i,i}} + \sum_{j \neq i}\exp^{w \cdot z^{cls}_{j,i}}}\right)
    \label{EQ:contrastive_loss_go}
\end{equation}
where $z_{j,i}^{cls}$ is the multi-modal fusion output for ``[CLS]'' token taking $x_j$ and $y_i$ as inputs; $w \in \mathcal{R}^{1 \times d}$ is the parameter in a linear layer\footnote{for clarity, we omit the bias term in the formula}. Based on~\eqref{EQ:contrastive_loss_go}, we optimize all parameters in our model $\theta = \{\theta_v, \theta_t, \theta_m\}$ in collaboration with~\eqref{EQ:contrastive_loss_ge} and~\eqref{EQ:contrastive_loss_te}.

In~\eqref{EQ:contrastive_loss_go}, a practical challenge is that we can hardly use all $(K-1)$ negative samples in the mini-batch, due to the high computational and memory cost in the multi-modal fusion. The $O(d(m+n)^2)$ complexity of self-attention layer makes it intractable to pass all $K\times K$ pairs into the multi-modal layers. Previous work solved this by performing random sampling to cut the number of negative samples to $K'$. However, randomly choosing negative samples may result in sub-optimal learning since the pairs are scarce. We therefore introduce a cascade sampling strategy to find hard negatives instead of random ones.

\xhdr{Cascade hard negative sampling}. To reduce the computational cost in~\eqref{EQ:contrastive_loss_go}, we choose among all possible video-text pairs a small subset which are most difficult. However, computing the alignment scores for all pairs using~\eqref{EQ:contrastive_loss_go} and then select the hard negatives is a ``chicken-and-egg'' problem. Instead, we propose to use the similarities between all video-text pairs computed in~\eqref{EQ:contrastive_loss_ge} and~\eqref{EQ:contrastive_loss_te} as the guidance. Specifically, for each text-video pair $(v_j, t_i)$, we take their global similarity $\bar{x}_j\cdot \bar{y}_i$ computed in ~\eqref{EQ:contrastive_loss_ge} and token-level similarity by aggregating $\sum_{p \in P_i} s(x_j, y^p_i)$ for all tokens of interest in $t_i$. Then we sum the two similarities as the alignment score for the given pair. For each text, we choose the top $K'$ aligned negative videos and vice versa. The resulting $2K \times (K'+1)$ pairs are then fed into the multi-modal fusion layers. Through this strategy, we can effectively select the difficult negative samples on the fly at no extra cost. Since the multi-modal fusion layers has more capacity (parameters) to distinguish these hard negatives from positive ones, our sampling strategy naturally prompts the cooperation between the three contrastive losses.

Finally, we present a comprehensive comparison to differentiate our model with previous works with respect to the used contrastive learning method in Table~\ref{Table:COntrastiveComparison}. 
\begin{table}[t]\footnotesize
\footnotesize
\resizebox{\linewidth}{!}{
\centering
\setlength{\tabcolsep}{3.0pt}{
\begin{tabular}{l c c c c c}
\toprule
Method & Token-aware & Early stage & Later stage &  Cascade & Loss\\
\midrule
VideoBert~\cite{videobert} & \xmark & \xmark & \cmark & \xmark & BCE\\
CBT~\cite{sun2019learning} & \xmark & \xmark & \cmark & \xmark & NCE \\
TJVE~\cite{Miech2019}   & \xmark & \cmark & \xmark & \xmark & Margin\\
MIL-NCE~\cite{Miech2020}    & \xmark & \cmark & \xmark & \xmark & NCE \\
ActBert~\cite{Zhu2020}    & \xmark & \xmark & \cmark & \xmark & BCE \\
UniVL~\cite{Luo2020}  & \xmark & \cmark & \cmark & \xmark & NCE \\
MMT~\cite{gabeur2020multi}  & \xmark & \cmark  & \xmark & \xmark & Margin \\
\oursloss{}(Ours)                & \cmark & \cmark & \cmark & \cmark & NCE \\
\bottomrule
\end{tabular}}}
\vspace{-7pt}
\caption{A comparison of video-language pretraining methods regarding contrastive learning strategies. ``Early stage'' and ``Later stage'' mean computing the loss before and after multi-modal fusion, respectively. ``Cascade'' means using cascade hard negative sampling.}
\vspace{-5pt}
\label{Table:COntrastiveComparison}
\end{table}

\subsection{Objective}

The training objective in our method is finding optimal $\theta=\{\theta_v, \theta_t, \theta_m\}$ by minimizing the combination of the above three contrastive losses:
\begin{equation}
\small
    \arg\min_{\theta_v, \theta_t, \theta_m} \sum_{i=1}^{N} \left(L_1 + \lambda_t L_2 + L_3\right)
    \label{EQ:combined_contrastive_loss}
\end{equation}
where $\lambda_t$ is the weight of token-level loss (0.5 by default). During inference, we make the prediction by summing the alignment scores from all the three scoring functions. 
\section{Experimental setup}
\label{Sec:experiment}
\subsection{Datasets}
In our experiments, we train and evaluate our model on the following established benchmarks:

\noindent\textbullet~\textbf{YouCook2}~\cite{Zhou2017a} consists of 2k videos about routine cooking activities of 89 recipes. Each video contains multiple video clips annotated with text descriptions by human annotators. Following~\cite{Miech2019,Miech2020}, we train our models on the training split, and report the text-video retrieval performance on around 3.5k validation clips. 

\noindent\textbullet~\textbf{MSR-VTT}~\cite{Xu2016c} contains 10k video clips associated with 200k sentences. There are two validation splits used in previous work. In~\cite{liu2019use,gabeur2020multi}, the training set has 9k clip-text pairs with the remaining 1k pairs for evaluation, which we denote by \textit{split1}. In~\cite{yu2018joint, Miech2019,Miech2020}, 1k clip-text pairs are sampled from the 3k pairs in test set for evaluation, while the original 7k pairs are used for training. We denote this by \textit{split2}. We report text-video retrieval results using both splits.

\noindent\textbullet~\textbf{ActivityNet}~\cite{krishna2017dense}. It consists of 20K YouTube videos, each of which is associated with multiple human-annotated captions. Following~\cite{zhang2018cross,gabeur2020multi}, we concatenate all the captions for a video into a paragraph and evaluate the paragraph-video retrieval on the ``val1'' split.

\noindent\textbullet~\textbf{Howto100M}~\cite{Miech2019}. 
We compare with previous work under the pretraining protocol on Howto100M~\cite{Miech2019, Miech2020, Zhu2020, Luo2020}. It was collected from YouTube and contains over 1.2M narrated videos associated with automatically generated transcripts. Each video contains over 100 clips on average.

To further verify the transferrability or our learned multi-modal representation from Howto100M, we also evaluate the action step localization and action segmentation on CrossTask~\cite{zhukov2019cross} and COIN~\cite{tang2019coin}, respectively.

\subsection{Settings}
\label{subsec:settings}
\vspace{-3pt}
Previous work use a variety of different video and language representations which we find significantly affect the final performance. We summarize different choices below:

\noindent\textbullet~\textbf{Video representations}. For 2D CNN, Resnet-152~\cite{He2015} is used to extract feature map and then globally pooled to 2048-d~\cite{Miech2019, Luo2020}. For 3D features, commonly used models are I3D~\cite{carreira2017quo}, R(2+1)D~\cite{tran2018closer} and S3D~\cite{xie2018rethinking}.
In~\cite{Zhu2020}, the authors further extract objects from the video clips. In~\cite{liu2019use, gabeur2020multi}, the authors use collaborative experts to extract features from audio, scene, OCR, face, speech, etc.

\noindent\textbullet~\textbf{Language representations}. There are primarily four variants: 1) GoogleNews pretrained word2vec (w2v)~\cite{w2v} used in~\cite{liu2019use, Miech2019, Miech2020}; 2) LSTM or Bidirectional LSTM~\cite{hochreiter1997long}; 3) pretrained BERT~\cite{Devlin2018} used in~\cite{videobert, Zhu2020,Luo2020,gabeur2020multi} and 4) OpenAI-GPT~\cite{radford2018improving} used in~\cite{liu2019use}.

In this paper, we use a pretrained BERT-base model for language representation as in~\cite{Zhu2020,Luo2020}. For video features, following~\cite{Miech2019, Miech2020, Luo2020}, we extract 2D CNN features using Resnet-152 (R-152) pretrained on ImageNet~\cite{deng2009imagenet}. For 3D CNN features, we use I3D (with Resnext-101 backbone) pretrained on Kinetics-400~\cite{Kay2017} and 
S3D~\cite{xie2018rethinking} pretrained on Howto100M~\cite{Miech2020}. The off-the-shelf pretrained weights are provided by~\cite{hara3dcnns} and~\cite{Miech2020}. For simplicity, we denote them by I3D-X101 and S3D-HM in the following.

Another discrepancy among different methods is the number of self-attention layers used in the model. In~\cite{Zhu2020}, the authors use 12 multi-modal self-attention layers while 6 video encoder layers and 2 multi-modal fusion layers are used in~\cite{Luo2020}. Differently, 4 multi-modal self-attention layers are used in~\cite{gabeur2020multi}. In this paper, for all our ablation studies below, we use 1 and 2 self-attention layers for our video encoder and multi-modal fusion, respectively. To compare with previous work on specific dataset, we use 2 video encoding layers. While pretraining the model with large-scale dataset Howto100M~\cite{Miech2019}, we increase to 4 video encoding layers for comparable model capacity to previous works~\cite{Zhu2020,Luo2020,gabeur2020multi}. Note that this largest model is still smaller than or on par with the aforementioned methods.

\begin{table}[t]
\centering
\resizebox{\linewidth}{!}{
\setlength{\tabcolsep}{3.0pt}{
    \begin{tabular}{@{}l@{\hspace{2pt}}l@{\hspace{7pt}}
    c@{\hspace{2pt}}c@{\hspace{2pt}}c@{\hspace{2pt}}c@{\hspace{7pt}}
    c@{\hspace{2pt}}c@{\hspace{2pt}}c@{\hspace{2pt}}c@{}}
    \toprule
    & & \multicolumn{4}{c}{YouCook2} & \multicolumn{4}{c}{MSR-VTT (\textit{split1})} \\
    \cmidrule{3-6}\cmidrule{7-10}   
    & Video Representation & R1$\uparrow$ & R5$\uparrow$ & R10$\uparrow$ & MR$\downarrow$ & R1$\uparrow$ & R5$\uparrow$ & R10$\uparrow$ & MR$\downarrow$ \\
    \toprule
    & R-152, Baseline    & 4.1 & 13.2 & 19.4 & 81.0 & 16.4 & 42.6 & 55.8 & 8.0 \\
    & R-152, Ours    & \textbf{4.6} & \textbf{14.1} & \textbf{20.4} & \textbf{71.0} & \textbf{18.9} & \textbf{46.2} & \textbf{58.8} &
    \textbf{7.0}\\
    \arrayrulecolor{black!20}\cmidrule{1-10}
    & I3D-X101, Baseline  & 2.1 & 8.1 & 12.7 & 125.0 & 14.7 & 40.83 & 53.2 & 9.0 \\
    & I3D-X101, Ours         & \textbf{2.6} & \textbf{8.9} & \textbf{13.2} & \textbf{115.0} & \textbf{20.6} & \textbf{44.0} & \textbf{56.9} & \textbf{7.0}\\   
    \arrayrulecolor{black!20}\cmidrule{1-10}
    & R-152+I3D-X101, Baseline & 4.2 & 13.5 & 20.0 & 75.0 & 16.6 & 45.4 & 58.5 & 7.0 \\    
    & R-152+I3D-X101, Ours & \textbf{4.7} & \textbf{14.3} & \textbf{21.9} & \textbf{68.0} & \textbf{23.1} & \textbf{50.5} & \textbf{64.0} & \textbf{5.0}\\    
    \arrayrulecolor{black!20}\cmidrule{1-10}
    & S3D-HM, Baseline & {13.8} & {37.2} & {51.1} & {10.0} & 18.7 & 47.2 & {62.2} & {6.0} \\
    & S3D-HM, Ours    & \textbf{16.1} & \textbf{40.3} & \textbf{52.2} & \textbf{9.0} & \textbf{23.9} & \textbf{51.4} & \textbf{65.0} & \textbf{5.0} \\   
    \arrayrulecolor{black!20}\cmidrule{1-10}
    & R-152+S3D-HM, Baseline & 13.3 & 35.8 & 48.9 & 11.0 & {21.4} & {48.1} & 61.5 & {6.0} \\    
    & R-152+S3D-HM, Ours & \textbf{15.8} & \textbf{39.8} & \textbf{52.4} & \textbf{10.0} & \textbf{24.5} & \textbf{52.8} & \textbf{65.5} & \textbf{5.0} \\
    \arrayrulecolor{black}\bottomrule
    \end{tabular}}
    }
    \vspace{-3pt}
    \caption{Text-video retrieval performance on YouCook2 and MSR-VTT with different feature types. S3D pretrained on HowTo100M outperforms others with large margin.}
    \label{tab:VideoRepresentation}
\end{table}

\subsection{Implementation details}
\vspace{-3pt}
For YouCook2 and MSR-VTT, the maximum number of video and text tokens are set to 48 and 30, respectively. For paragraph-video retrieval on ActivityNet, we set them both to 256. The 2D R-152 feature is extracted for one frame per second, and then globally pooled to 2048-d. For 3D CNN features, we follow~\cite{Miech2019} to sample video frames at 24 fps and extract an I3D-X101 feature every 16 frames. This results in 1.5 2048-d feature per second. For Eq.~\ref{EQ:contrastive_loss_ge} and~\ref{EQ:contrastive_loss_te}, we set the temperatures $\tau_1$ and $\tau_2$ both equal to 1. 

\xhdr{Training on separate datasets}. In this setting, we train models from scratch using the training set provided in YouCook2,  MSR-VTT and ActivityNet separately.
We train the model for 30k iterations with batch size 128. For each training sample, we use our cascade sampling strategy to sample 8 hard negatives. We use Adam~\cite{kingma2014adam} as the optimizer with initial learning rate $1e^{-4}$. A linear learning rate decay is applied after 5k warm-up iterations. The weight decay is set to $1e^{-5}$.

\xhdr{Pretraining and finetuning}. We pretrain our model on Howto100M~\cite{Miech2019}. Since the original annotated video clips in Howto100M are usually short with a few seconds, we merge the adjacent clips so that the resulted text has at least 10 words. We use Adam~\cite{kingma2014adam} as the optimizer with initial learning rate $1e^{-4}$. We train the model for 500k iterations with batch size 64, and also sample 8 hard negatives for each sample using our cascade sampling strategy. After pretraining, we finetune the pretrained models on different datasets using the same setting as above except for a lower initial learning rate $2e^{-5}$ and less finetuning iterations 20k.

\xhdr{Evaluation metrics}. For text-video retrieval, we use Recalls at different points (Recall@n or Rn, with n as a specific number) and Median Rank (MR) as the metrics following previous works~\cite{Zhu2020,Luo2020}. In all tables, we use $\uparrow$ or $\downarrow$ to indicate higher or lower is better, respectively. 


\section{Results}
\vspace{-3pt}
We first evaluate text-video retrieval performance and then study whether the learned representations can be transferred to other tasks on CrossTask and COIN.

\subsection{Text-video retrieval}
\vspace{-3pt}
\subsubsection{Comparing with baselines}
\vspace{-3pt}
We first show the comparisons with baselines to inspect the effects of different components in our model.

\begin{table}
\resizebox{\linewidth}{!}{
\setlength{\tabcolsep}{3.0pt}{
\centering
    \begin{tabular}{@{}l@{\hspace{1pt}}c@{\hspace{8pt}}
    c@{\hspace{3pt}}c@{\hspace{3pt}}c@{\hspace{3pt}}c@{\hspace{8pt}}
    c@{\hspace{3pt}}c@{\hspace{3pt}}c@{\hspace{3pt}}c@{}}
    \toprule
    &  & \multicolumn{4}{c}{YouCook2} & \multicolumn{4}{c}{MSR-VTT (\textit{split1})} \\
    \cmidrule{3-6}\cmidrule{7-10}   
    Losses & Cascade & R1$\uparrow$ & R5$\uparrow$ & R10$\uparrow$ & MR$\downarrow$ & R1$\uparrow$ & R5$\uparrow$ & R10$\uparrow$ & MR$\downarrow$ \\
    \midrule
    $L_1$       & n/a  & 14.1 & 35.7 & 48.8 & 11.0 & 22.9 & 49.7 & 61.7 & 6.0 \\
    $L_3$      & n/a   & 13.3 & 35.8 & 48.9 & 11.0 & 21.4 & 48.1 & 61.5 & 6.0  \\    
    $L_1+L_3$     & \xmark & 13.9 & 37.4 & 50.7 & 10.0 & 22.5 & 50.8 & 64.1 & 5.0 \\ 
    $L_1+L_3$      & \cmark & 15.0 & 38.7 & 51.3 & 10.0 & 23.7 & 51.3 & 63.9 & 5.0 \\
    $L_1+L_2+L_3$ & \cmark & \textbf{15.8} & \textbf{39.8} & \textbf{52.4} & \textbf{10.0} & \textbf{24.5} & \textbf{52.8} & \textbf{65.5} & {5.0} \\    
    \bottomrule
    \end{tabular}}
    }
    \vspace{-7pt}
    \caption{Text-video retrieval performance with different technique ensembles. It shows that using our proposed two techniques produce best results. All experiments use R-152+S3D-HM video features.}
    \label{tab:AlignmentType}
\end{table}

\begin{table}
\resizebox{\linewidth}{!}{
\setlength{\tabcolsep}{3.0pt}{
\centering
    \begin{tabular}{@{}l@{\hspace{8pt}}
    c@{\hspace{3pt}}c@{\hspace{3pt}}c@{\hspace{3pt}}c@{\hspace{8pt}}
    c@{\hspace{3pt}}c@{\hspace{3pt}}c@{\hspace{3pt}}c@{}}
    \toprule
    &  \multicolumn{4}{c}{YouCook2} & \multicolumn{4}{c}{MSR-VTT (\textit{split1})} \\
    \cmidrule{2-5}\cmidrule{6-9}   
    Token of Interest & R1$\uparrow$ & R5$\uparrow$ & R10$\uparrow$ & MR$\downarrow$ & R1$\uparrow$ & R5$\uparrow$ & R10$\uparrow$ & MR$\downarrow$ \\
    \midrule
    None      & 15.0 & 38.7 & 51.3 & 10.0 & 23.7 & 51.3 & 63.9 & 5.0 \\
    det+adp   & 14.7 & 38.5 & 51.2 & 10.0 & 23.3 & 51.0 & 63.5 & 5.0  \\    
    noun      & 15.4 & 39.3 & 51.8 & 10.0 & 24.0 & 51.8 & 65.1 & 5.0 \\ 
    verb      & 15.3 & 39.0 & 51.4 & 10.0 & 23.9 & 52.1 & 64.8 & 5.0 \\
    noun+verb & \textbf{15.8} & \textbf{39.8} & \textbf{52.4} & {10.0} & \textbf{24.5} & \textbf{52.8} & \textbf{65.5} & {5.0} \\    
    \bottomrule
    \end{tabular}}
    }
    \vspace{-7pt}
    \caption{Text-video retrieval performance with different tokens of interest for computing token-level contrastive loss. ``det'' means determiner; ``adp'' means adposition. We use the same video features as in Table~\ref{tab:AlignmentType}.}
    \vspace{-5pt}
    \label{tab:TOI}
\end{table}

\xhdr{Video representations}. We train our model with different video representations as described above and compare it with the baseline model which has identical architecture but merely trained with $L_3$ as depicted in \eqref{EQ:contrastive_loss_go}. The baseline contrastive learning method has been adopted in a number of previous works~\cite{Zhu2020,Luo2020}. This comparison can verify the effectiveness of our proposed contrastive learning method considering two models have exactly the same number of parameters. In Table~\ref{tab:VideoRepresentation}, we can see our proposed method outperforms baseline across all feature types introduced in Sec.~\ref{subsec:settings} on both YouCook2 and MSR-VTT. Note that our model uses exactly the same number of parameters to the baseline model. These consistent improvements demonstrate the effectiveness and generalization ability of our proposed method. As mentioned above, we also observe the text-video retrieval performance significantly depends on the feature types. We can find 3D features (I3D-X101 and S3D-HM) in general outperform 2D feature (R-152), which is expected since 2D feature does not capture the motions in the videos. Among all three feature types, S3D-HM outperforms the other two with large margin, which demonstrates the potential to learn good video representation by pretraining on large-scale noisy dataset (Howto100M~\cite{Miech2019}). Because Howto100M mainly contains instructional videos, it is more close to YouCook2 than MSR-VTT, and hence we see more gain on YouCook2. These comparisons indicate video representations matter much to the final performance.

\begin{table}[t]
\resizebox{\linewidth}{!}{
\setlength{\tabcolsep}{3.2pt}
\centering
    \begin{tabular}{@{}lllccccccccccccc@{}}
    \toprule
    \multirow{2}{*}{Model}     &  \multirow{2}{*}{Lang.} &  \multirow{2}{*}{Video} & \multicolumn{4}{c}{YouCook2} \\
    \cmidrule{4-7}\cmidrule{9-12}\cmidrule{13-16}
         &  &  & R1$\uparrow$ & R5$\uparrow$ & R10$\uparrow$ & MR$\downarrow$ \\    
    \midrule
    Random & -- & -- & 0.0 & 0.2 & 0.3 & 1675 \\
    TVJE~\cite{Miech2019} &  w2v  & R-152+I3D-X101  & 4.2 & 13.7 & 21.5 & 65\\
    UniVL(v1)~\cite{Luo2020}   &  BERT & R-152+I3D-X101   & 3.4   & 10.8   & 17.8   & 76 \\    
    \oursloss{} (Ours) &  BERT &  R-152+I3D-X101 & \textbf{4.9} & \textbf{14.7} & \textbf{21.7} & \textbf{63} \\      
    \arrayrulecolor{black!20}\cmidrule{1-7}
    UniVL(v3)~\cite{Luo2020}   &  BERT & S3D-HM  & 7.7 &  23.9 & 34.7 & 21 \\     
    \oursloss{} (Ours) &  BERT & S3D-HM & \textbf{16.6} & \textbf{40.3} & \textbf{53.1} & \textbf{9.0} \\       
    \arrayrulecolor{black}
    \bottomrule
    \end{tabular}}
    \vspace{-7pt}
    \caption{Comparing text-video retrieval on YouCook2.}
    \label{tab:youcook2}
\end{table}

\begin{table}[t]
\resizebox{\linewidth}{!}{
\setlength{\tabcolsep}{3.2pt}
\centering
    \begin{tabular}{@{}lllcccccccccccccc@{}}
    \toprule
    \multirow{2}{*}{Model}     &  \multirow{2}{*}{Lang.} & \multirow{2}{*}{Video} & \multicolumn{4}{c}{MSR-VTT}  \\
    \cmidrule{4-8}\cmidrule{9-12}
         &  &  & R1$\uparrow$ & R5$\uparrow$ & R10$\uparrow$ & MR$\downarrow$ \\    
    \midrule
    Random & -- & -- & 0.1 & 0.5 & 1.0 & 500.0  \\
    JSFusion~\cite{yu2018joint} & BiLSTM & R-152 & 10.2 & 31.2 & 43.2 & 13.0 \\
    JPoSE~\cite{wray2019fine} & w2v & TSN+Flow & 14.3 & 38.1 & 53.0 & 9.0 \\
    TVJE~\cite{Miech2019} &  w2v  & R-152+I-101 & {12.1} & {35.0} & {48.0} & 12.0 \\
    UniVL(v1)$^*$~\cite{Luo2020}   &  BERT & R-152+I-101 & 14.6   & 39.0   & 52.6   & 10.0 \\       
    \oursloss{} (Ours) &  BERT &  R-152+I-101 & \textbf{19.2} & \textbf{44.7} & \textbf{57.2} & \textbf{7.0} \\     
    \arrayrulecolor{black!20}\cmidrule{1-7}
    CE~\cite{liu2019use} & GPT & Collaborative Experts & 20.9 & 48.8 & 62.4 & 6.0 \\    
    MMT~\cite{gabeur2020multi} & BERT & Collaborative Experts & 24.6 & 54.0 & 67.1 & \textbf{4.0} \\
    \oursloss{} (Ours) &  BERT &  R-152+S3D-HM & \textbf{26.7} & \textbf{54.5} & \textbf{68.2} & \textbf{4.0} \\  
    \arrayrulecolor{black}
    \bottomrule
    \end{tabular}}
    \vspace{-7pt}
    \caption{Comparing text-video retrieval on MSR-VTT. The upper block and bottom block use \textit{split2} and \textit{split1}, respectively. We report them separately for fair comparison.}
    \label{tab:msrvtt}
\end{table}

\begin{table}
\resizebox{\linewidth}{!}{
\setlength{\tabcolsep}{3.2pt}
\centering
    \begin{tabular}{@{}lllcccccccccccccc@{}}
    \toprule
    \multirow{2}{*}{Model}     &  \multirow{2}{*}{Lang.} & \multirow{2}{*}{Video} & \multicolumn{4}{c}{ActivityNet} \\
    \cmidrule{4-8}\cmidrule{9-12}\cmidrule{13-16}
         &  & & R1$\uparrow$ & R5$\uparrow$ & R10$\uparrow$ & MR$\downarrow$ \\    
    \midrule
    Random & - & - & 0.02 & 0.1 & 1.02 & 2458 \\
    DenseCap~\cite{krishna2017dense} & LSTM & C3D & 14.0 & 32.0 & 65.0 & 34 \\
    FSE~\cite{zhang2018cross} & GRU & C3D+TSN-Inception & 18.2 & 44.8 & 89.1 & 7.0 \\
    CE~\cite{liu2019use} & GPT & Collaborative Experts & 18.2 & 47.7 & 91.4 & 6.0 \\
    MMT~\cite{gabeur2020multi} & BERT &  Collaborative Experts & 22.7 & 54.2 & 93.2 & 5.0 \\
    \oursloss{} (Ours) &  BERT &  R-152+S3D-HM & \textbf{25.8} & \textbf{56.3} & \textbf{93.8} & \textbf{4.0} \\       
    \arrayrulecolor{black}
    \bottomrule
    \end{tabular}}
    \vspace{-7pt}
    \caption{Comparing text-video retrieval on ActivityNet.}
    \label{tab:activitynet}
\end{table}

\xhdr{Component Analysis}. In our method, we combine $L_1$, $L_2$, and $L_3$ during training and inference. Here, we study how they perform separately and contribute to the final performance. In Table~\ref{tab:VideoRepresentation}, we use R-152+S3D-HM as the video feature and report the results with different loss combinations. As we can see, solely using $L_1$ (row 1) or $L_2$ (row 2) for contrastive learning results in sub-optimal video-text alignment. Simply combining them together (row 3) improves the performance on two datasets. This implies that different levels of contrastive learning can be complementary to each other, which supports our earlier hypothesis that these two losses are synergistic with each other for a better video-text alignment. When incorporating the hard negative mining via our cascade sampling strategy (row 4), it further improves the performance. Finally, we can see adding token-level contrastive loss $L_3$ can further improve the performance across all settings~(row 5).

\xhdr{Tokens of Interest}. We further study the effect of different tokens of interest on the model performance. By default, our model uses the noun and verb as the tokens of interest to compute the token-level contrast loss. Here, we vary them to other types such as adposition (adp) and determiner (det) for investigation. In Table~\ref{tab:TOI}, we replace ``noun+verb'' with ``det+adp'', ``noun'' and ``verb'' and report the numbers on two text-video retrieval datasets. As we can see, using ``det+adp'' as the target tokens is worse than the baseline without any token-level contrastive loss. ``noun'' and ``verb'' can both improve the performance while ``noun'' is slightly better than ``verb''. Finally, combining noun and verb together achieves the best performance. These results align with our intuition to use nouns and verbs as the target token for fine-grained alignment between texts and videos considering they are usually grounded to video contents.




\subsubsection{Comparing with state-of-the-art}
We compare with previous works under three protocols: 1) training and evaluating on separate datasets; 2) pretraining on Howto100M and evaluating zero-shot performance and 3) finetuning pretrained model on separate datasets.


\begin{table}[t]
\centering
\begin{minipage}{0.98\linewidth}
\centering
  \includegraphics[width=0.98\linewidth]{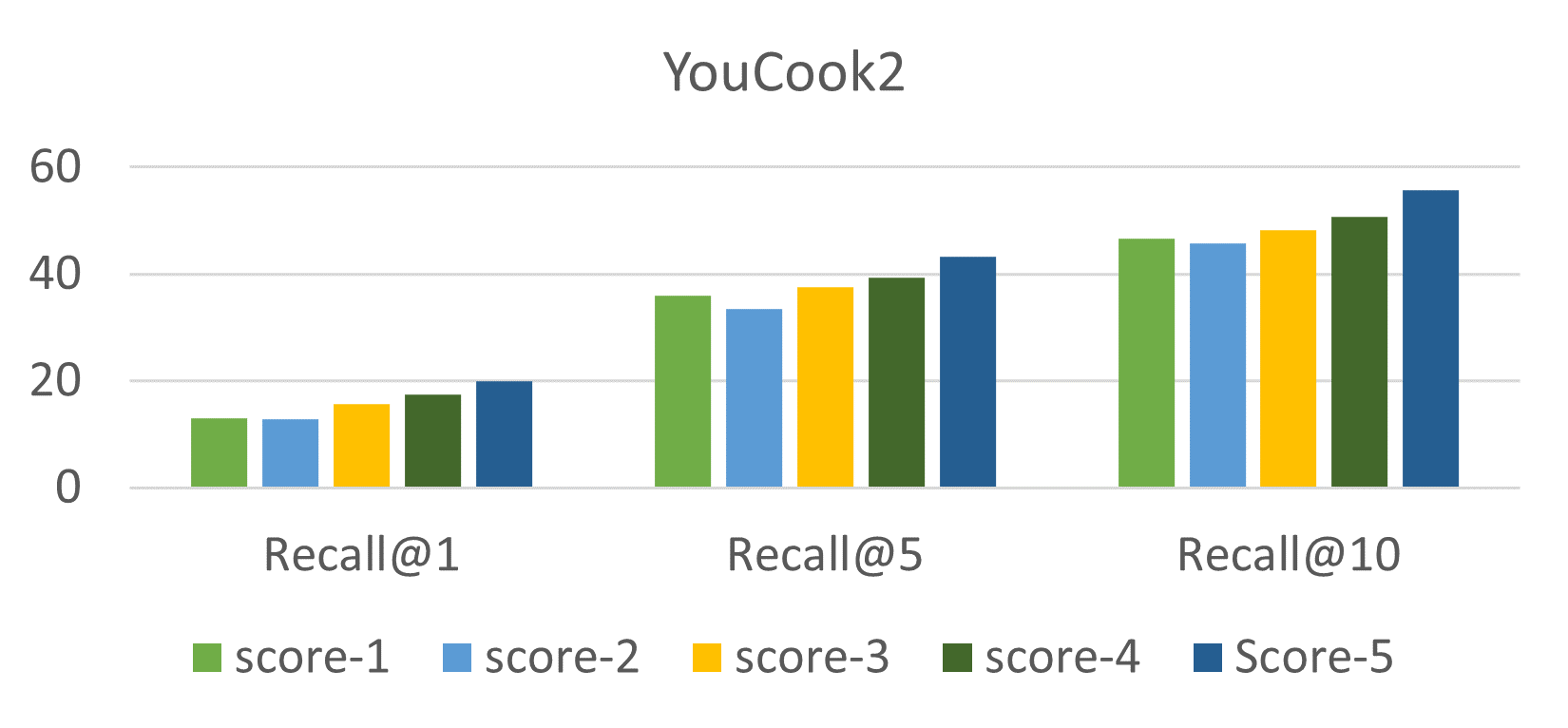}
  \vspace{-10pt}
  \includegraphics[width=0.98\linewidth]{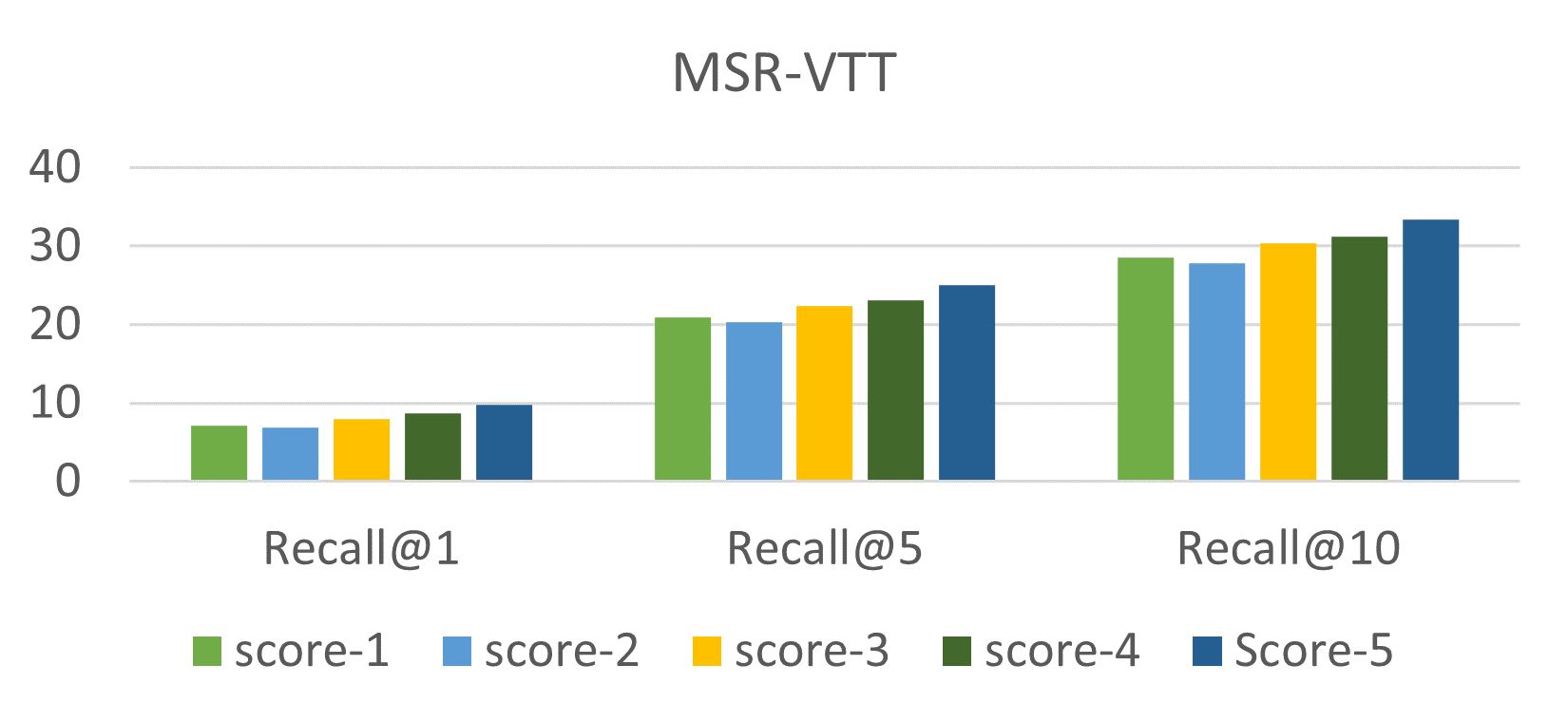}
  \captionof{figure}{Zero-shot performance on YouCook2 and MSR-VTT for different settings. score-{1-5} correspond to the five settings in Table~\ref{tab:AlignmentType} from top to bottom.}
\label{fig:zero-shot-ablations}
\end{minipage}
\vspace{-10pt}
\end{table}

\begin{table*}[t]
\begin{minipage}{0.7\linewidth}
\resizebox{\linewidth}{!}{
\setlength{\tabcolsep}{3.3pt}
\centering
    \begin{tabular}{@{}l@{\hspace{7pt}}l@{\hspace{7pt}}l
    @{\hspace{5pt}}c@{\hspace{5pt}}c
    @{\hspace{5pt}}c@{\hspace{5pt}}c@{\hspace{5pt}}c
    @{\hspace{5pt}}c@{\hspace{5pt}}c@{\hspace{5pt}}c@{\hspace{5pt}}c
    @{\hspace{5pt}}c@{\hspace{5pt}}c@{\hspace{5pt}}c@{\hspace{5pt}}c@{}}
    \toprule
    & \multirow{2}{*}{Model}
    & \multirow{2}{*}{Video} 
    & \multicolumn{4}{@{}c@{}}{YouCook2} 
    & \multicolumn{4}{@{}c@{}}{MSR-VTT} &  \multicolumn{4}{@{}c@{}}{ActivityNet} \\
    \cmidrule{4-16}
         & & & R1$\uparrow$ & R5$\uparrow$ & R10$\uparrow$ & MR$\downarrow$ & R1$\uparrow$ & R5$\uparrow$ & R10$\uparrow$ & MR$\downarrow$ & R1$\uparrow$ & R5$\uparrow$ & R50$\uparrow$ & MR$\downarrow$\\    
    \arrayrulecolor{black}\midrule
    \multirow{5}{*}{\rotatebox{90}{Zero-shot}} 
    & TJVE~\cite{Miech2019}    & R-152+I-101 & 6.1           & 17.3          & 24.8          & 46.0         & 7.5          & 21.2          & 29.6           & 38.0 & -- & -- & -- & -- \\
    & ActBERT~\cite{Zhu2020}         & O-101+ R(2+\!1\!)D & 9.6   & 26.7 & {38.0} & {19.0} & {8.6} & {23.4} & {33.1} & {36.0} & -- & -- & -- & --\\    
    & MIL-NCE~\cite{Miech2020}  & S3D-HM & 15.1          & {38.0} &  {51.2} & {10.0} & \textbf{9.9} & {24.0} & {32.4} & {29.5} & -- & -- & -- & -- \\  
    & \oursloss{} (Ours)        & S3D-HM & \textbf{19.9} & \textbf{43.2} & \textbf{55.7} & \textbf{8.0} & {9.8} & \textbf{25.0} & \textbf{33.4} & \textbf{29.0} & -- & -- & -- & --\\   
    \arrayrulecolor{black}\midrule
    \multirow{6}{*}{\rotatebox{90}{Finetuned}} 
    & TJVE~\cite{Miech2019}  & R-152+I3D-X101 & 8.2     & 24.5 & 35.3 & 24.0 & 14.9          & 40.2          & 52.8          & 9.0 & -- & -- & -- & -- \\    
    & UniVL(v3)~\cite{Luo2020}   & S3D-HM & {28.9} & {57.6} & {70.0} & \textbf{4.0} & 21.2           & 49.6           & 63.1          &6.0 & -- & -- & -- & -- \\    
    & \oursloss{} (Ours)      & S3D-HM  &  \textbf{29.6}  &  \textbf{59.7} & \textbf{72.7} & \textbf{4.0} & \textbf{24.8} & \textbf{52.1} & \textbf{64.0} & \textbf{5.0} & 28.3 & 56.8 & 92.6 & 4.0 \\     
    \arrayrulecolor{black!20}\cmidrule{2-15}
    & MMT~\cite{gabeur2020multi} & Collaborative Experts & -- & -- & -- & -- & 26.6 & 57.1 & 69.6 & \textbf{4.0} & 28.7 & \textbf{61.4} & \textbf{94.5} & 3.3 \\
    & \oursloss{} (Ours)   & R-152+S3D-HM  &  27.3 & 56.5 & 68.8 & \textbf{4.0} & \textbf{28.4} & \textbf{57.8} & \textbf{71.2} & \textbf{4.0} & \textbf{30.4} & 61.2 & 93.4 & \textbf{3.0} \\ 
    \arrayrulecolor{black}\bottomrule
    \end{tabular}
    }
    \vspace{-3pt}
    \caption{A complete comparison of \oursloss{} under zero-shot and finetuning evaluation protocols. Note that the zero-shot and upper part of finetuned performance for MSR-VTT is on \textit{split2}, while the bottom is on \textit{split1} for fair comparison.}
    \label{tab:pretraining}
\end{minipage}
\hspace{1em}
\begin{minipage}{0.28\linewidth}
\resizebox{\linewidth}{!}{
\centering
    \begin{tabular}{lcc}
    \toprule
    Method & CrossTask & COIN \\
    \midrule
    Alayrac~\textit{et al.}~\cite{alayrac2016unsupervised} & 13.3 & -- \\
    Zhukov~\textit{et al.}~\cite{zhukov2019cross} & 22.4 & -- \\
    Supervised~\cite{zhukov2019cross} & 31.6 & -- \\
    NN-Viterbi~\cite{richard2018neuralnetwork} & -- & 21.2\\
    CBT~\cite{sun2019learning} & -- & 53.9 \\
    TVJE~\cite{Miech2019} & 33.6 & -- \\
    MIL-NCE~\cite{Miech2020} & 40.5 & 61.0 \\
    ActBert~\cite{Zhu2020} & 41.4 & 57.0 \\
    UniVL(v3)~\cite{Luo2020} & 42.0 & \textbf{70.0} \\
    \midrule
    \oursloss{}~(Ours) & \textbf{42.5} & 68.4 \\
    \bottomrule
    \end{tabular}
    }
    \vspace{-3pt}
    \caption{Action step localization on CrossTask (avg. recall) and action segmentation on COIN (acc.).}
    \label{tab:othertasks}
\end{minipage}
\end{table*}

\xhdr{Results on separate datasets}. We separately show the comparisons on YouCook2, MSR-VTT and ActivityNet in Table~\ref{tab:youcook2}, \ref{tab:msrvtt} and~\ref{tab:activitynet}. For a fair comparison with previous works, we use the same or similar features as listed in the tables. As we can see, our method outperforms all previous work across all datasets. These results validates its effectiveness to learn video-text alignment. Note that previous works either use a variety of loss functions~\cite{Luo2020, li2020hero} or a collection of multiple features~\cite{liu2019use, gabeur2020multi}. In contrast, we achieve the best performance using a \textit{simpler} contrastive learning pipeline with smaller model size. This supports our earlier claim on the efficiency. Comparing the numbers in Table~\ref{tab:VideoRepresentation}, Table~\ref{tab:youcook2} and Table~\ref{tab:msrvtt}, we can find our model achieves better performance with the same video features when using deeper video encoder (2 layers \textit{v.s.} 1 layer).

\xhdr{Zero-shot and finetuned performance}. In Table~\ref{tab:pretraining}, we show the comparisons across different models pretrained on Howto100M. In the upper part of the table, we compare the zero-shot performance on YouCook2 and MSR-VTT. We do not evaluate on ActivityNet since it has different number of input video tokens compared with the pretrained model and thus is not directly compatible to the pretrained model. As we can see, \oursloss{} outperforms previous works significantly on YouCook2 and slightly on MSR-VTT. Since YouCook2 has closer domain gap to Howto100M than MSR-VTT, the improvement brought by large-scale pretraining is more significant. However, on MSR-VTT, our model still outperforms MIL-NCE~\cite{Miech2020} which uses the same video features. In Fig.~\ref{fig:zero-shot-ablations}, we show the zero-shot performance on YouCook2 and MSR-VTT when pretraining our models with different contrastive losses as listed in Table~\ref{tab:AlignmentType}. Accordingly, it shows our proposed contrastive losses gradually improve the performance, and combining all techniques achieves the best performance. Based on the pretrained model, we further finetune it on specific datasets. In our experiments, we use two feature S3D-HM and R-152+S3D-HM, to compare with the methods with the same/similar settings. As we can see, our model using S3D-HM outperforms UniVL~\cite{Luo2020} using the same feature but more video encoder layers (6). Different from zero-shot results, we observe more improvement on MSR-VTT than YouCook2 after finetuning. This implies that finetuning on specific datasets can compensate the domain gap to the pretraining datasets. To compare with the methods using features extracted from collaborative experts~\cite{gabeur2020multi}, we enrich our video representation by adding 2D R-152 feature, which achieves better performance on MSR-VTT, and better Recall@1 and Median Rank on ActivityNet. Note that this combination hurts the performance on YouCook2, and we witnessed a similar trend for models without pretraining in Table~\ref{tab:VideoRepresentation}. Finally, comparing with the results without pretraining in Table~\ref{tab:youcook2},~\ref{tab:msrvtt} and~\ref{tab:activitynet}, we clearly find large-scale pretraining and finetuning brings substantial improvements consistently.

\subsection{Other video-related tasks}
\vspace{-2pt}
Following~\cite{Miech2019,Zhu2020,Luo2020}, we evaluate action step localization performance on CrossTask dataset~\cite{zhukov2019cross}. It covers 18 tasks and each video contains multiple video segments annotated with action steps and natural language descriptions. Similar to~\cite{Miech2019,Zhu2020,Luo2020}, we use our model to compute the similarity between each frame and the action step descriptions, which results in a score matrix. Using the official algorithm provided by~\cite{zhukov2019cross}, we can find the optimal frame-wise order of action steps for a video. By comparing it with the ground-truth annotations, we compute the recall for each task and then do the average. According to the results in Table~\ref{tab:othertasks}, our model achieves the best performance compared with previous works. 
This indicates that our model can learn good video-language representations.


We further evaluate our pretrained model on action segmentation task on COIN dataset, following~\cite{Miech2020, Zhu2020}. Unlike the above task, action segmentation does not rely on texts, and thus can be used to evaluate the learned video representation. As shown in Table~\ref{tab:othertasks}, our method significantly outperforms MIL-NCE and ActBert, and achieves comparable performance to UniVL. This indicates that our model is also a good video representation learner.

\section{Conclusion}
\vspace{-3pt}
In this paper, we introduced \oursloss{}, a simple yet effective contrastive learning method for learning video-text alignment. It is aimed at addressing two existing issues in current contrastive learning pipelines: missing fine-grained alignment and inefficient sampling for multi-modal fusion. Without introducing any extra parameters, our method achieved promising results on three text-video retrieval benchmarks under various evaluation protocols. We further demonstrated the learned representations can be effectively transferred to other tasks such as action step localization and segmentation. Based on all these encouraging results, we believe \oursloss{} is a good alternative to conventional contrastive learning pipeline.


{\small
\bibliographystyle{ieee_fullname}
\bibliography{egbib}
}
\appendix
\section{Tokens of interest}
\begin{figure*}[t]
    \centering
    \includegraphics[width=\linewidth, height=0.2\linewidth]{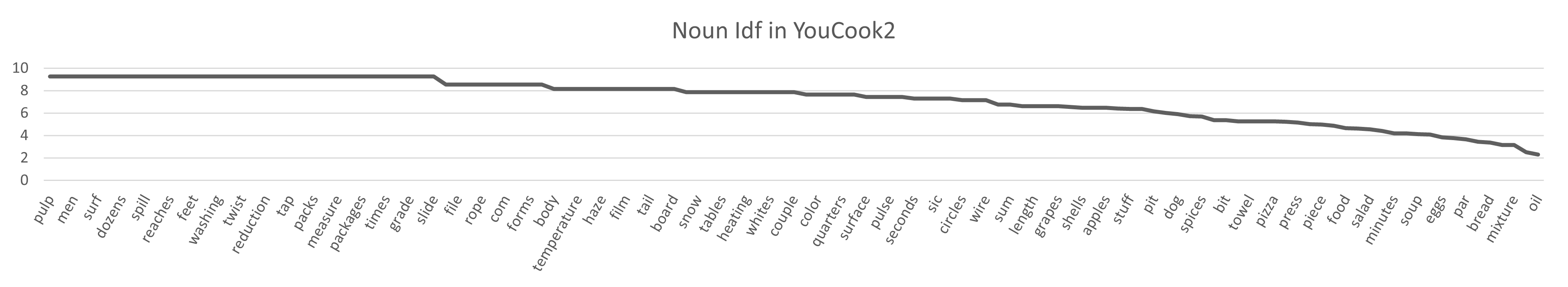}
    \vspace{-5pt}
    \includegraphics[width=\linewidth, height=0.2\linewidth]{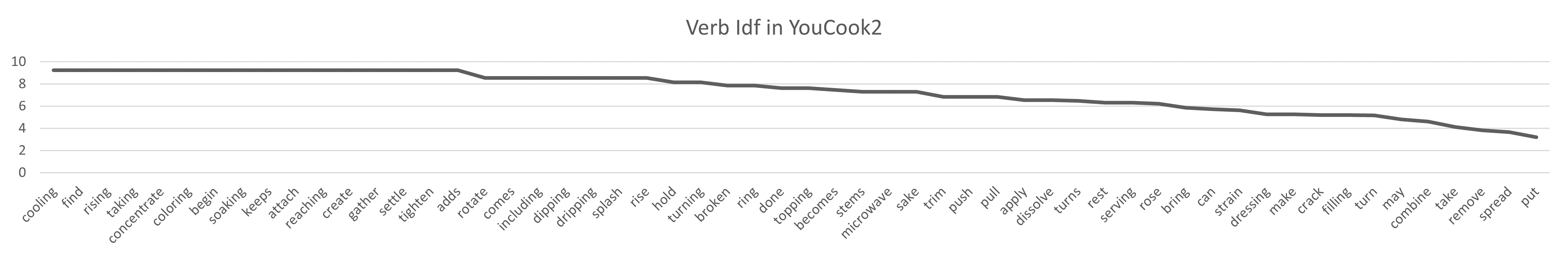}
    \vspace{-5pt}
    \includegraphics[width=\linewidth, height=0.2\linewidth]{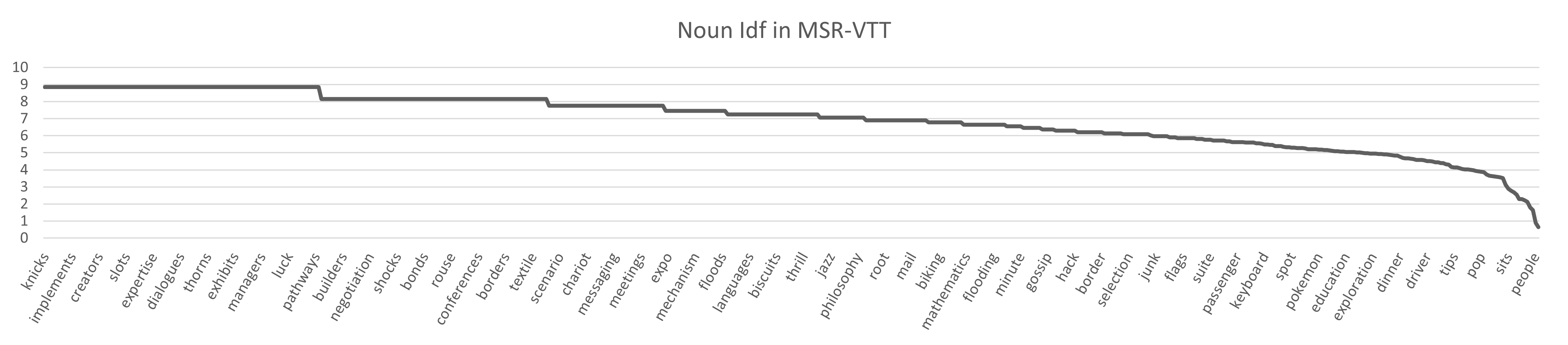}
    \vspace{-5pt}
    \includegraphics[width=\linewidth, height=0.2\linewidth]{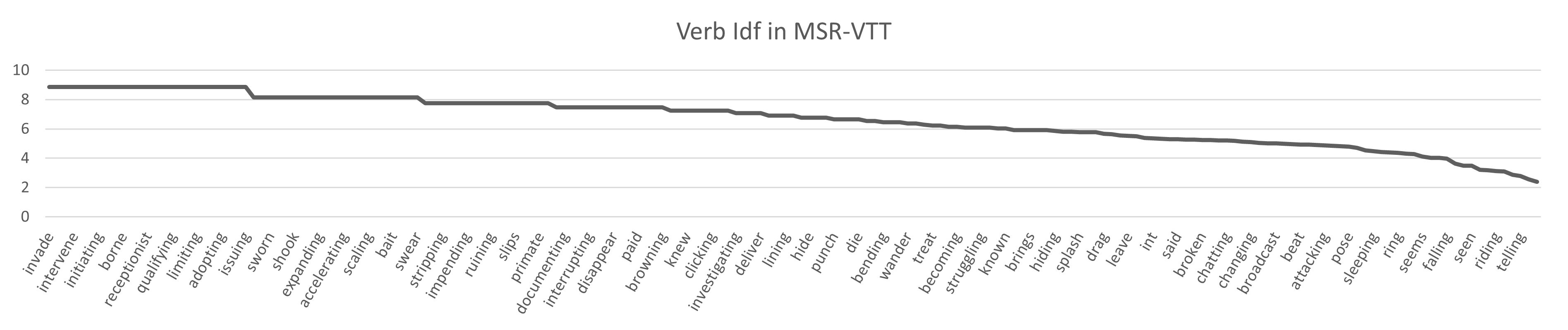}    
    \vspace{-5pt}
    \includegraphics[width=\linewidth, height=0.2\linewidth]{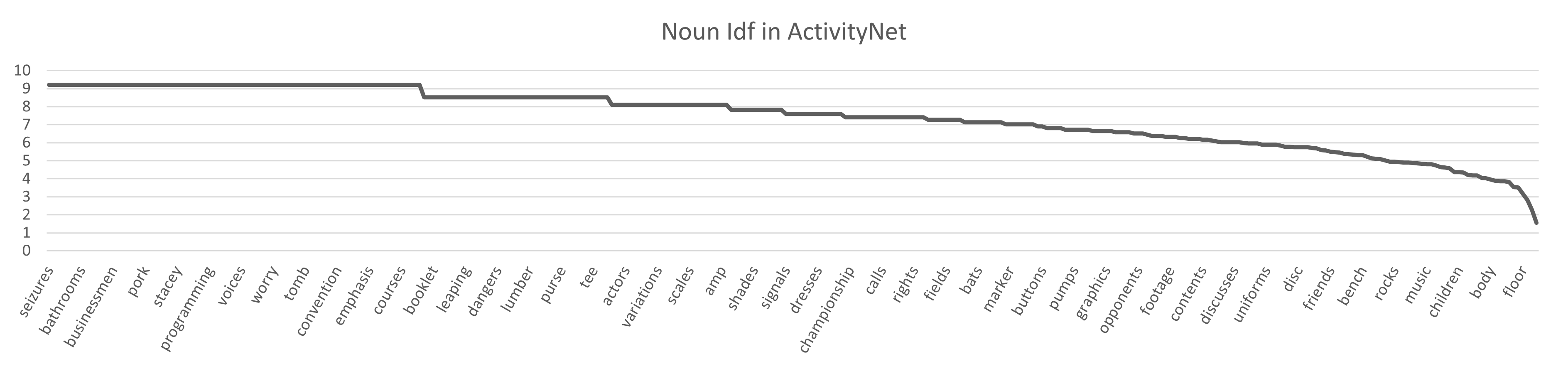}
    \vspace{-5pt}
    \includegraphics[width=\linewidth, height=0.2\linewidth]{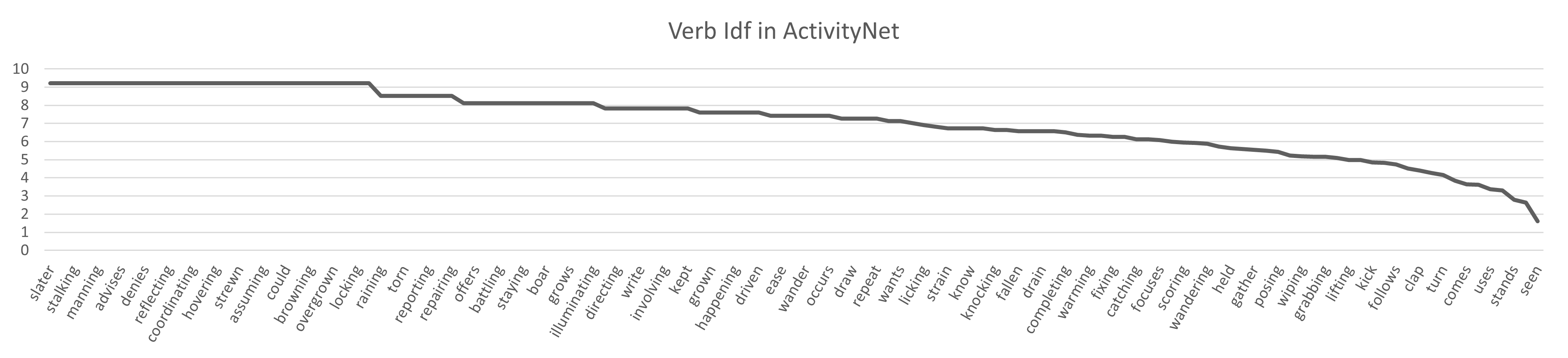}        
    \vspace{-20pt}
    \caption{Token inverse document frequency (IDF) for noun and verb in YouCook2 and MSR-VTT. For clarity, we evenly sample the tokens and show their IDFs. From left to right, the noun/verb becomes more and more frequent gradually.}
    \vspace{-10pt}
    \label{fig:token_idf}
\end{figure*}

\vspace{-5pt}
\begin{table}[h]
\centering
    \begin{tabular}{lccc}
    \toprule
    Dataset & Noun & Verb & All\\
    \midrule
    YouCook2~\cite{Zhou2017a} & 378 & 168 & 2,144\\
    MSR-VTT~\cite{Xu2016c} & 4,415 & 1,463 & 15,740\\
    ActivityNet~\cite{krishna2017dense} & 2,602 & 1,021 & 9,059 \\
    \bottomrule
    \end{tabular}
    \vspace{-10pt}
    \caption{Token statistics for each dataset.}
    \label{tab:token_statistics}
    \vspace{-10pt}
\end{table}
We extract tokens of interest (T.O.I) using the pos-tagger provided by Spacy~\cite{spacy2}. In Table~\ref{tab:token_statistics}, we show the statistics of tokens for three datasets. For each token that is tagged at \textit{VERB} or \textit{NOUN}, we compute the inverse document frequency (idf) by:
\begin{equation}
\small
    idf(token) = \log \frac{|D|}{1 + |\{d \in D: token \in d\}|}
    \label{EQ:idf}
\end{equation}
where $D$ is the full set of corpus, which are the captions in the training set for a dataset; the denominator counts the number of captions which contain a specific token. Based on~\eqref{EQ:idf}, we can compute the idf for each token of interest. The smaller the idf, the more frequent it appears in the corpus. We do not compute the tf term since usually a token only appears once in a single sentence. The full list of tokens and corresponding idfs can be found in Fig.~\ref{fig:token_idf}. For a given sentence, we first assign the computed idfs to its nouns and verbs and then normalize the idfs, which are then used to weigh the token-level contrastive losses.




\section{Contribution of three contrastive losses}
\vspace{-5pt}
\begin{table}[!ht]
\centering
\footnotesize
\resizebox{0.9\linewidth}{!}{
    \begin{tabular}{lcccc}
    \toprule
    Loss & R@1 & R@5 & R@10 & MR \\
    \midrule
    Early stage only        & 14.1 & 35.7 & 48.8 & 11.0 \\
    Later stage only        & 13.3 & 35.8 & 48.9 & 11.0 \\    
    \midrule
    Early stage        & 15.3 & 39.3 & 51.9 & \textbf{10.0} \\
    Token-level        & 15.0 & 39.5 & 51.4 & 11.0 \\
    Later stage        & 14.3 & 38.4 & 50.6 & 11.0 \\
    Fused              & \textbf{15.8} & \textbf{39.8 }& \textbf{52.4} & \textbf{10.0} \\
    \bottomrule
    \end{tabular}}
    \vspace{-10pt}
    \caption{Text-video retrieval performance using separate alignment scores on YouCook2.}
    \label{tab:sep_performance}
    \vspace{-10pt}
\end{table}

In this part, we investigate the contributions of three contrastive losses used in our model. After we train the video-text alignment model using all three losses, we report the performance using separate alignment scores in Table~\ref{tab:sep_performance}. For reference, the top two rows are the performance for using early stage only and later stage only contrastive learning to train the model. The bottom four rows are the separate performance at different stages for our model. As we can see, combining three contrastive losses during training can boost the performance for both early and later stage (row 3 \textit{v.s.} row 1, row 5 \textit{v.s.} row 2). This indicates that the three losses are synergistic to each other for a better video-text alignment. On the other hand, the early stage alignment achieves better performance than other two (token-level and later stage), while the fused score is the best. We suspect that this is because early stage alignment is trained with all text-video pairs at sentence-level. In contrast, token-level contrast focuses on single tokens and the multi-modal fusion layers merely see a small part of hard text-video pairs.

\section{Effect of cascade sampling}
\vspace{-5pt}
The proposed cascade sampling helps the later stage contrastive learning to focus on hard negative samples. As shown in our main submission, adding cascade sampling will improve the performance. We suspect this is because cascade sampling helps learn a better later stage alignment. To verify this, we compare the later stage alignment across three different settings: 1) merely applying later stage contrastive loss; 2) combine early state and later stage contrastive losses and 3) using cascade sampling for later stage contrastive loss. We report the results on YouCook2 in Table~\ref{tab:cascade_sampling}. Here, note that we only use the later stage alignment scores for evaluating the performance. As we can see, combining early stage and later stage together slightly improves the performance. This is probably because early stage contrastive loss helps to learn a better video and language encoder, from which the multi-modal module takes better representations for cross-modal fusion. After applying the cascade sampling for the later stage contrastive loss, the performance is further improved. Since our cascade sampling strategy can send more difficult samples to the later stage, the cross-modal fusion layers can learn more discriminative representations for video-text alignment. These results validate that the hard negative mining through cascade sampling indeed helps to improve the later-stage text-video alignment, and hence the final performance.

\begin{table}[!ht]
\centering
\resizebox{0.98\linewidth}{!}{
    \begin{tabular}{lcccc}
    \toprule
    Setting & R@1 & R@5 & R@10 & MR \\
    \midrule
    Later stage only   & 13.3 & 35.7 & 48.8 & 11.0  \\
    Early stage + Later stage & 13.6 & 35.9 & 49.1 & 11.0 \\
    Cascade sampling   & \textbf{14.5} & \textbf{38.3} & \textbf{50.7} & 11.0  \\
    \bottomrule
    \end{tabular}}
    \vspace{-10pt}
    \caption{Text-video retrieval performance on YouCook2 only using later stage alignment score for different settings.}
    \vspace{-10pt}
    \label{tab:cascade_sampling}
\end{table}

\section{Effect of video encoder layers}
\vspace{-5pt}
In our main paper, we noticed the number of video encoder layers affects the final performance. To have a more comprehensive study, we use R-152 and S3D-HM as the 2D and 3D features and train the video-text alignment model on YouCook2 with different video encoder layers. As shown in Table~\ref{tab:performance_with_depth}, using more video encoder layers can significantly boost the text-video retrieval performance. Particularly, when no video encoder layers are used, the model can hardy capture the long-range temporal dynamics, and thus performs poorly. Once we add one video encoder layer, the performance improves significantly. With the increase of encoder layers, the performance is further improved, which is reasonable since more video encoder layers can encode more complicated video contents and dynamics.

\begin{table}[!ht]
\setlength{\tabcolsep}{1.3pt}
\vspace{-5pt}
\centering
\resizebox{0.93\linewidth}{!}{
    \begin{tabular}{ccccccc}
    \toprule
    \multirow{2}{*}{\makecell{\#video \\ enc. layers}}  & \multirow{2}{*}{\makecell{\#params. \\ (M)}} & \multirow{2}{*}{\makecell{FLOPs \\ (G)}} &  \multicolumn{4}{c}{YouCook2} \\
    &  &  & R@1 & R@5 & R@10 & MR \\
    \midrule
    0   & 126.5 &   3.86 & 14.0 & 35.7 & 49.5 & 11.0  \\
    1   & 133.6 &   4.11 & 15.8 & 39.8 & 52.4 & 10.0  \\
    2   & 140.7 &   4.45     & 15.9 & \textbf{40.5} & 53.8 & \textbf{9.0}   \\
    4   & 154.9 &   5.14  & \textbf{16.4} & \textbf{40.5} & \textbf{54.3} & \textbf{9.0}   \\
    \bottomrule
    \end{tabular}}
    \vspace{-10pt}
    \caption{Text-video retrieval performance on YouCook2 with different video encoder layers using R-152+S3D-HM.}       
    \label{tab:performance_with_depth}
    \vspace{-10pt}
\end{table}




\section{Comparing model size and FLOPs}
\vspace{-5pt}
Finally, we attempt to compare the model sizes and computational costs for different methods. Unfortunately, all previous methods did not report FLOPs and only MMT~\cite{gabeur2020multi} discussed \#params. However, the results in Table~\ref{tab:performance_with_depth} imply that bigger model can usually achieve better performance. Therefore, it is necessary to have a comparison of model size and computational cost between our model and those from other methods. For other methods which do not report the numbers, we estimate them based on the descriptions in the original paper. Table~\ref{tab:model_size} summarizes the comparisons and also reports the \#params and FLOPs (all underlined numbers are estimated based on the descriptions in original papers). As shown, our largest model has comparable size and FLOPs to others.

\begin{table}[!ht]
\centering
\footnotesize
\resizebox{0.9\linewidth}{!}{
\setlength{\tabcolsep}{1.5pt}
\vspace{-5pt}
\begin{tabular}{lcccccc}
    \toprule
    \multirow{2}{*}{method} & \multirow{2}{*}{text} & \multirow{2}{*}{video} &  \multicolumn{2}{c}{mm} &  \multirow{2}{*}{\makecell{\#params.\\ (M)}} & \multirow{2}{*}{\makecell{FLOPs \\ (G)}} \\
    \cmidrule{4-5}
    & & & self & cross & &\\
    \midrule
    ActBert~\cite{Zhu2020} & 12 & 12 & 0 & 24 & \underline{369.1} & \underline{13.80} \\
    MMT~\cite{gabeur2020multi}  & 12 & 4  & 0 & 0  & 133.3 & \underline{4.63} \\
    UniVL~\cite{Luo2020}   & 12 & 6  & 2 & 0  & \underline{169.0} & \underline{5.82} \\
    Our largest  & 12 & 4 & 2  & 0  & 154.9 & 5.14 \\
    \bottomrule
    \end{tabular}}
    \vspace{-10pt}
    \caption{Comparison of model size and FLOPs. ``mm'' means multi-modal fusion, and ``self'' means self-attention layers while ``cross'' means cross-modal attention.}
    \label{tab:model_size}
    \vspace{-10pt}
\end{table}

\section{Visualizations}
\begin{figure*}[t]
    \includegraphics[width=\linewidth]{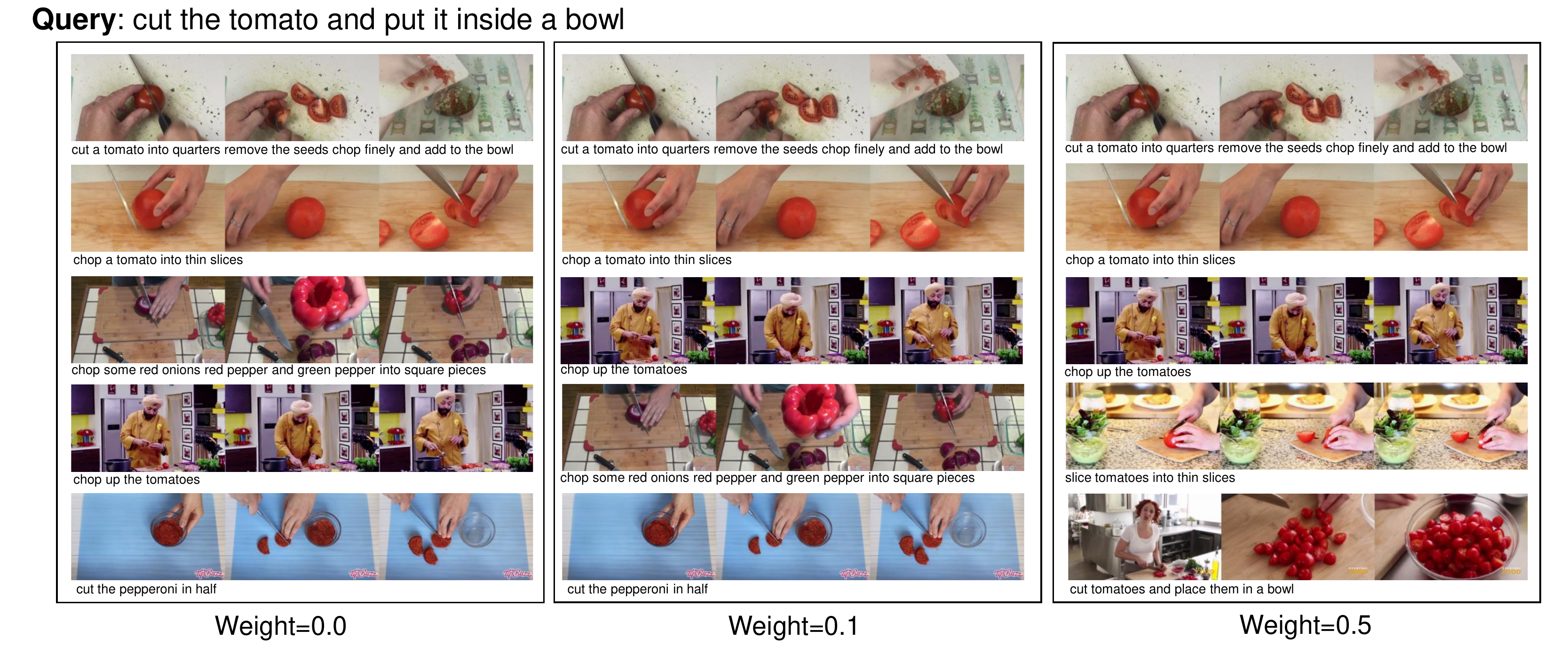}
    \includegraphics[width=\linewidth]{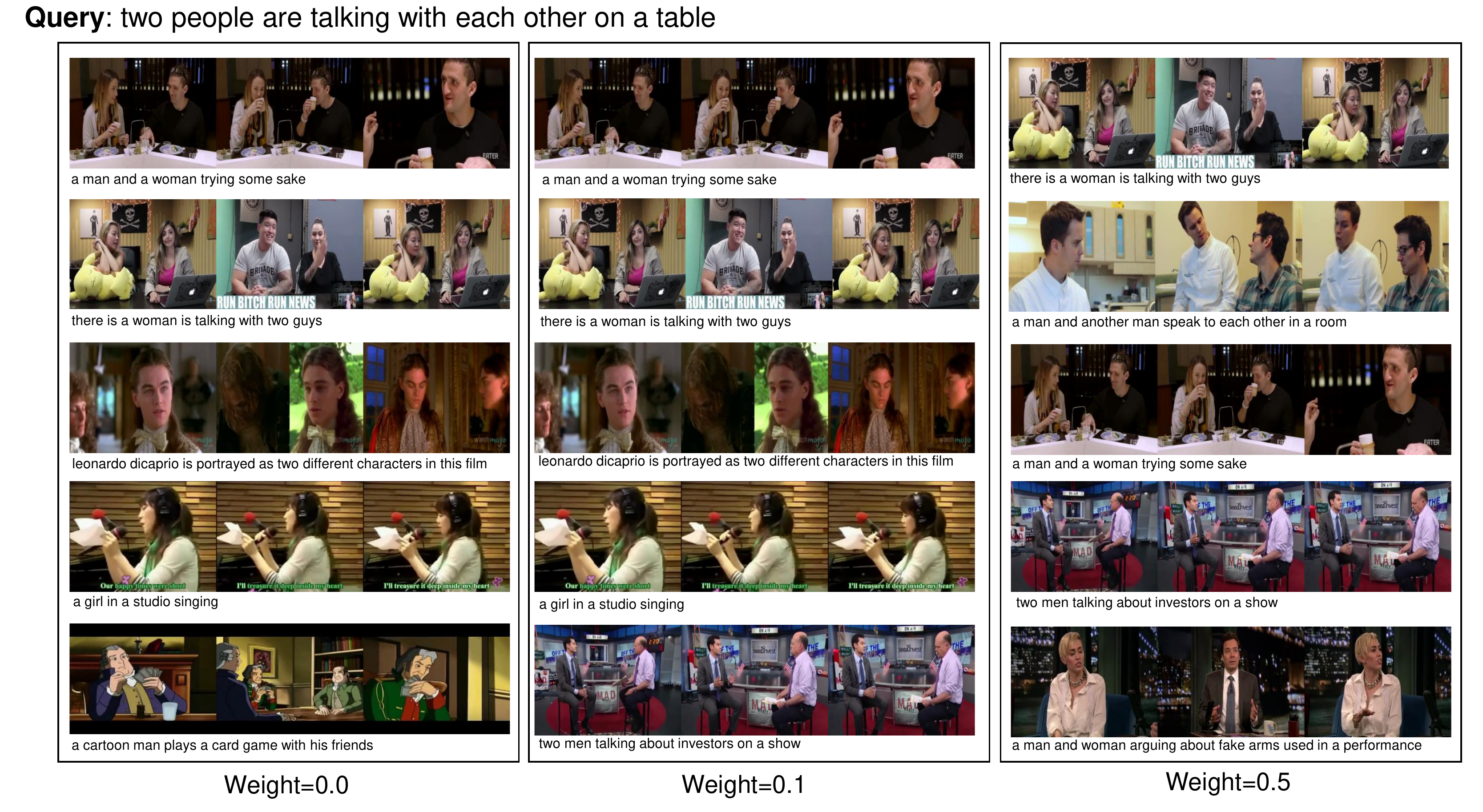}
    \vspace{-20pt}
    \caption{Text-video retrieval results given a query on YouCook2 (top) and MSR-VTT (bottom). In each block, we show top 5 ranked videos from top to bottom. From left to right, we gradually increase the token-level alignment weight from 0.0 to 0.1 and then 0.5 (default in our main experiments). The change of the top 5 results demonstrate the benefit of token-level contrast when performing text-video retrieval. Below each video (depicted by three side-by-side frames), we show the associated descriptions provided in the original dataset. Better viewed by enlarging the figure.}
    \label{fig:visualization}
\end{figure*}

\vspace{-5pt}
We visualize the text-video retrieval results by varying the weights for the token-level alignment scores during testing. In Fig.~\ref{fig:visualization}, we show two text-video retrieval examples on YouCook (top) and MSR-VTT (bottom). From top to bottom, the five rows in each block correspond to the top five retrieved results from the whole test set. As we can see, when we gradually increase the weight for the token-level alignment score, there are more related videos appearing in the top five candidates. For YouCook2, when we set the weight equal to 0.0, the third and fifth video are not well-aligned with the query since they are both not about ``tomato''. When we increase the weight to 0.1, we can observe the the fourth video moves to the third place. After we increase the weight to 0.5, we can see all top-5 videos are about cutting tomato. Similarly, for MSR-VTT, we can see the last three videos are not about ``two people talking on a table''. When we increase the weight to 0.1, the fifth video is replaced with a more matched video. Keeping increase the weight to 0.5, we can obtain the top 5 videos all about ``two people talking with each other on a table''. These visualizations demonstrate the efficacy of our proposed token-level contrastive learning. 

\end{document}